\documentclass[10pt,twocolumn,letterpaper]{article}
\usepackage{amsmath}
\usepackage{amssymb}
\usepackage{amsthm}
\usepackage{authblk}
\usepackage{latexsym}
\usepackage{array}
\usepackage{booktabs}
\usepackage{multirow}
\usepackage{makecell}
\usepackage{graphicx}
\usepackage{stfloats}
\usepackage[shortlabels, inline]{enumitem}
\usepackage{appendix}
\usepackage{color}
\usepackage[hyphens]{url}
\usepackage[ruled,linesnumbered]{algorithm2e}
\usepackage{cvpr}
\usepackage[pagebackref,breaklinks,colorlinks,allcolors=cvprblue]{hyperref}

\definecolor{cvprblue}{rgb}{0.21,0.49,0.74}

\newtheorem{corollary}{Corollary}
\newtheorem{theorem}{Theorem}
\newtheorem{definition}{Definition}

\newtheorem{assumption}{Assumption}

\def\paperID{342} 
\def\confName{CVPR}
\def\confYear{2025}

\title{Order-Robust Class Incremental Learning: Graph-Driven \\ Dynamic Similarity Grouping}
\author[1]{Guannan Lai$^{\dag}$}
\author[1, 2]{Yujie Li$^{\dag}$}
\author[1]{Xiangkun Wang}
\author[3]{Junbo Zhang}
\author[4]{Tianrui Li}
\author[1]{Xin Yang$^{\star}$}
\affil[1]{School of Computing and Artificial Intelligence, Southwestern University of Finance and Economics, 
\authorcr \textit{Email: aignlai@163.com, \{liyj1201, xiangkunwang18\}@gmail.com, yangxin@swufe.edu.cn}}

\affil[2]{LIACS, Leiden University,}

\affil[3]{JD Intelligent Cities Research, \textit{Email: msjunbozhang@outlook.com}}

\affil[4]{Southwest Jiaotong University, \textit{Email: trili@swjtu.edu.cn}}

\begin{document}
\maketitle

\footnotetext[1]{$^{\dag}$ Equal contribution, sorted alphabetically.}
\footnotetext[2]{$^{\star}$ Corresponding author.}

\begin{abstract}
Class Incremental Learning (CIL) aims to enable models to learn new classes sequentially while retaining knowledge of previous ones. Although current methods have alleviated catastrophic forgetting (CF), recent studies highlight that the performance of CIL models is highly sensitive to the order of class arrival, particularly when sequentially introduced classes exhibit high inter-class similarity. To address this critical yet understudied challenge of \textbf{class order sensitivity}, we first extend existing CIL frameworks through theoretical analysis, proving that grouping classes with lower pairwise similarity during incremental phases significantly improves model robustness to order variations. Building on this insight, we propose \textbf{G}raph-\textbf{D}riven \textbf{D}ynamic \textbf{S}imilarity \textbf{G}rouping (\textbf{GDDSG}), a novel method that employs graph coloring algorithms to dynamically partition classes into similarity-constrained groups. Each group trains an isolated CIL sub-model and constructs meta-features for class group identification. 
Experimental results demonstrate that our method effectively addresses the issue of class order sensitivity while achieving optimal performance in both model accuracy and anti-forgetting capability.
Our code is available at \url{https://github.com/AIGNLAI/GDDSG}.

\end{abstract}

\section{Introduction}
Class Incremental Learning (CIL) necessitates that the model dynamically acquires knowledge of new classes while preserving the knowledge of previously learned classes within an infinite sequence of tasks \cite{wang2019forward,ke2021achieving,NEURIPS2023_15294ba2}. 
CIL is realistic but a great challenge for deep neural networks \cite{parisi2019continual}, where existing works devoted to overcoming catastrophic forgetting (CF) and encouraging knowledge transfer across different tasks \cite{mccloskey1989catastrophic,ye2020heterogeneous,zhao2021mgsvf,wang2024comprehensive}.
With the rapid advancement of CIL, a growing number of methods \cite{yoon2019scalable,li2024hessian,shan2024order} have been introduced to address the problem of CF from the perspective of the order in which classes appear (or task order).
In practice, the arrival order of each class and the tasks to which they belong are random and the order in which tasks arrive is uncontrollable \cite{bell2022effect}, further resulting in \textit{Class order sensitivity} and \textit{Intra-task class conflicts} \cite{lin2023theory}. 
Therefore, designing an order-robust CIL method is essential for the community.

\begin{figure}[t]
    \centering
    \includegraphics[width=1\linewidth]{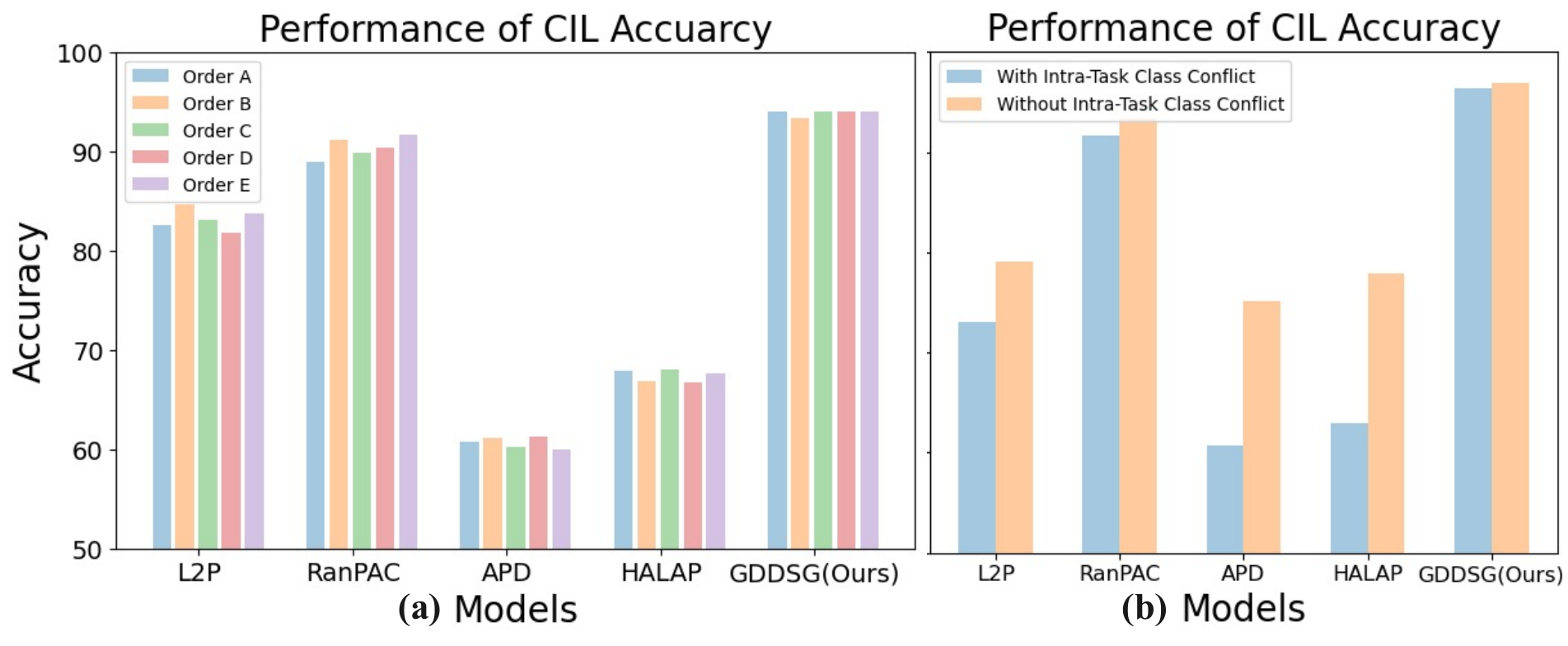}
    \caption{The crucial challenges of CIL (illustration on CIFAR100 dataset). On the left subfigure (a), each model’s performance is shown under varying class orders, testing its robustness to class order sensitivity. On the right subfigure (b), the model’s performance is shown when classes within the same task are similar, evaluating its resilience to intra-task classes with high similarities.}
    \label{fig: intro}
\end{figure}

\textit{Class order sensitivity} refers to the model exhibiting significant performance variations depending on the sequence in which classes are introduced \cite{shan2024order}. 
This phenomenon is prevalent in real-world applications (see \autoref{fig: intro}(a)). 
For instance, in online recommendation systems, the order in which user data classes are received at different time points is difficult to control. If the system initially receives data from relatively few classes, the introduction of subsequent classes may impair the system’s adaptability, resulting in unstable model performance on new tasks.
Furthermore, the model’s parameters may be overfitted to the classes of early tasks, diminishing its ability to generalize to subsequent task with new classes.
Although existing research, such as APD \cite{yoon2019scalable} and HALRP \cite{li2024hessian}, have attempted to mitigate the class order sensitivity problem by modifying network structures, their effectiveness remains limited and has not fundamentally addressed this challenge. 
Thus, designing a model capable of maintaining stable performance across varying class orders remains a critical unsolved issue in CIL.

\textit{Intra-task class conflicts} refers to the discrepancies in model performance caused by the similarity between classes that are trained simultaneously in a specific task (see \autoref{fig: intro}(b)). 
In real-world applications, where the arrival of classes in the data stream is uncontrollable, significant similarities among classes can severely impact the model's resilience. 
For example, in a specific task from a sequence of tasks, a model may be trained to recognize different breeds within the same species. Within this task, due to the high similarity of features across categories, the model needs to develop resilience in distinguishing between closely related classes.
However, existing CIL methods struggle to address this challenge, primarily due to the inherent limitations of the task setting. As CIL incrementally processes different classes, it cannot globally account for all class information, causing class conflicts to accumulate during training and negatively impact model performance. Thus, alleviating class conflicts and improving the model’s generalization ability remains a significant challenge in CIL.

Hence, to tackle the challenges of class order sensitivity and Intra-task class similarity sensitivity, we first conduct an in-depth analysis beyond existing theories. 
Our theoretical findings suggest that as class similarity decreases in CIL, the model's robustness to class order increases, which, in turn, mitigates knowledge conflicts both across different tasks and within individual tasks.
Then, we propose a similarity graph-based dynamic grouping method, called \textbf{Graph-Driven Dynamic Similarity Grouping (GDDSG)}, to maintain the centroids of existing classes and dynamically groups tasks based on class similarity, assigning classes with lower similarity to the same group.
This approach innovatively organizes class groups in CIL by utilizing a graph-based technique to minimize inter-group similarity. It dynamically assigns classes based on adaptive similarity thresholds and optimal graph coloring, thereby enhancing model robustness and computational efficiency across tasks.
In the incremental learning process, GDDSG continuously updates existing groups or creates new ones, training a separate model for each group. Consequently, during the prediction phase, decisions are made by aggregating the outputs of multiple models.

Hence, our contributions can be summarised as follows:
\begin{itemize}
    \item In this paper, we elaborate on existing theories and derive an important Corollary: when the similarity between classes is low, the model's sensitivity to class order is significantly reduced, leading to a decrease in class conflicts.

    \item Then, we provide a detailed introduction to the proposed GDDSG method, including its foundational algorithms and basic processes.

    \item Additionally, we conduct extensive comparative experiments to validate the effectiveness of GDDSG, highlighting its advantages and potential in incremental learning tasks.
\end{itemize}
\section{Related Work}

\textbf{Class-Incremental Learning (CIL)} necessitates a model that can continuously learn new classes while retaining knowledge of previously learned ones \cite{dohare2024loss,cao2023retentive,zhou2023revisiting,zhou2024continual}, which can be roughly divided into several categories. 
Regularization-based methods incorporate explicit regularization terms into the loss function to balance the weights assigned to new and old tasks \cite{kirkpatrick2017overcoming,aljundi2018memory,wang2022continual,li2017learning}.
Replay-based methods address the problem of catastrophic forgetting by replaying data from previous classes during the training of new ones. This can be achieved by either directly using complete data from old classes \cite{lopez2017gradient,riemer2018learning,cha2021co2l,wang2022foster} or by generating samples \cite{shin2017continual,zhu2022self}, such as employing GANs to synthesize samples from previous classes \cite{cong2020gan,liu2020generative}.
Dynamic network methods adapt to new classes by adjusting the network structure, such as adding neurons or layers, to maintain sensitivity to previously learned knowledge while acquiring new tasks. This approach allows the model's capacity to expand based on task requirements, improving its ability to manage knowledge accumulation in CIL \cite{wang2022coscl,wang2023incorporating,aljundi2017expert,ostapenko2021continual}.
Recently, CIL methods based on pre-trained models (PTMs)
\cite{cao2023retentive,chen2022adaptformer,zhou2024continual} have demonstrated promising results.
Prompt-based methods utilize prompt tuning \cite{jia2022visual} to facilitate lightweight updates to PTMs. By keeping the pre-trained weights frozen, these methods preserve the generalizability of PTMs, thereby mitigating the forgetting in CIL \cite{smith2023coda,wang2022dualprompt,NEURIPS2023_d9f8b5ab,wang2022learning,wang2022dualprompt,smith2023coda,li2024learning}. 
Model mixture-based methods mitigate forgetting by saving models during training and integrating them through model ensemble or model merge techniques \cite{gao2023unified,wang2023isolation,wang2024hierarchical,zheng2023preventing,zhou2023learning,zhou2024expandable}.
Prototype-based methods draw from the concept of representation learning \cite{ye2017learning}, leveraging the robust representation capabilities of PTMs for classification with NCM classifiers \cite{panos2023first,zhou2023revisiting,mcdonnell2024ranpac}.

\textbf{The Order in CIL} remains a significant and unresolved challenge \cite{wang2024comprehensive}. APD \cite{yoon2019scalable} effectively addresses the problem of CF by decomposing model parameters into task-shared and sparse task-specific components, thereby enhancing the model's robustness to changes in class order. HALRP \cite{li2024hessian}, on the other hand, simulates parameter conversion in continuous tasks by applying low-rank approximation to task-adaptive parameters at each layer of the neural network, thereby improving the model's order robustness. However, the optimization strategies employed by these methods are confined to the network architecture itself and do not fundamentally resolve the underlying issues.
Recent theoretical analyses of CIL \cite{lin2023theory,shan2024order,bell2022effect,wu2021curriculum} indicate that prioritizing the learning of tasks (or classes) with lower similarity enhances the model's generalization and resistance to forgetting. Building on these theories, we conducted further research and developed corresponding methods in the following sections.

\section{Problem Formulation and Theory Analysis}

\subsection{Problem Formulation} 
\begin{definition} \textbf{(Class Incremental Learning (CIL))}
Given a sequence of tasks denoted as $1, ...,t,...$, each task $i$ is associated with a training set (i.e., ground-truth data) $\mathcal{D}^i = \{X^i, Y^i\}$, where $X^i$ represents the set of training samples and $Y^i$ is the set of labels. 
For task $i$, the set of classes is denoted as $CLS^i$ with the size of $|CLS^i|$, representing the number of classes in task $i$. 
With new tasks incrementally appearing, the goal of CIL is to learn a \emph{unified model} $\Phi: \mathcal{D}^i \to \mathbb{R}^d$ mapping input data to an embedding space equipped with a classifier $f(\cdot)$ that can perform well on all the tasks it has been learned.
\end{definition}

Note that for any pair of tasks $i$ and $j$ with $1 \leq i, j \leq n$ and $i \neq j$, the sets of classes $CLS^i$ and $CLS^j$ are disjoint and data from other tasks is unavailable at the current task, ensuring distinctiveness and non-overlapping nature between classes across each task. 

\subsection{The Effect of Class Ordering in CIL}

In \cite{lin2023theory}, the authors theoretically derived the expected forgetting value and expected generalization error for CIL under a linear model, where \( w_t^* \) denotes the optimal parameters of the model for the \( t \)-th task:
\begin{theorem}\label{Therom: ther} When $p \ge n + 2 $, we must have:
\begin{align}
    \mathbb{E}[F_T] &= \frac{1}{T-1}\sum_{i = 1}^{T-1}[(r^T-r^i)\lVert w_i^*\rVert^2+\sum_{j>i}^Tc_{i,j}\lVert w_j^* -  w_i^*\rVert^2 \nonumber 
    \\ &+\frac{p\sigma^2}{p-n-1}(r^i-r^T)],
    \label{eq: e_for}
\end{align}
\begin{align}
    \mathbb{E}[G_T] &= \frac{r^T}{T}\sum_{i = 1}^{T-1}\lVert w_i^*\rVert^2 + \frac{1-r}{T}\sum_{i = 1}^T r^{T-i}\sum_{k=1}^T \lVert w_k^* -  w_i^*\rVert^2 \nonumber 
    \\ &+ \frac{p\sigma^2}{p-n-1}(1-r^T).
    \label{eq: e_gen}
\end{align}
where the overparameterization ratio \( r = 1 - \frac{n}{p} \) in this context quantifies the degree of overparameterization in a model, where \( n \) represents the sample size, and \( p \) denotes the number of model parameters \cite{muthukumar2020harmless,hastie2022surprises}. The coefficients \( c_{i,j} = (1 - r)(r^{T-i} - r^{j-i} + r^{T-j}) \), with \( 1 \leq i < j \leq T \), correspond to the indices of tasks, and \( \sigma \) denotes a coefficient representing the model's noise level.
\end{theorem}

\autoref{Therom: ther} made a significant contribution to the study of class order in CIL, particularly in the two key expressions: \(\sum_{j>i}^T c_{i,j} \lVert w_j^* - w_i^* \rVert^2\) in \autoref{eq: e_for} and \(\sum_{i=1}^Tr^{T-i} \sum_{k=1}^T \lVert w_k^* - w_i^* \rVert^2\) in \autoref{eq: e_gen}. These formulas highlight the crucial role that class order plays in CIL. Building on this theory, further in this work, we derive sufficient conditions to ensure order robustness.

\begin{corollary}\label{Corollary: cor} A \textbf{sufficient condition} for the reduction of \(\mathrm{Var}(\mathbb{E}[G_T])\) and \(\mathrm{Var}(\mathbb{E}[F_T])\) is that the sum of the squared distances between the optimal parameters of tasks increases, i.e., \(\sum_{i,j=1}^T \lVert w_i^* - w_j^* \rVert^2\) becomes larger.
\end{corollary}

\autoref{Corollary: cor} integrates the similarity between tasks with the model's robustness to class order. Through \autoref{eq: e_for} and \autoref{eq: e_gen}, we observe that both forgetting and generalization errors are influenced by the optimal model gap between any two tasks, represented by \(\lVert w^*_i - w^*_j \rVert^2\) for tasks \(i\) and \(j\). This gap serves as a measure of task similarity: the smaller the gap, the greater the similarity.
\autoref{Corollary: cor} demonstrates that a smaller similarity between tasks enhances the model's robustness in terms of generalization and resistance to forgetting across different class orders. This finding offers valuable insights for the design of new methods.
%\textit{\textbf{The proof of \autoref{Corollary: cor} can be found in the supplementary material.}}
\section{The Proposed Method: GDDSG}\label{sec4}
\begin{figure*}
    \centering
    \includegraphics[width=\linewidth]{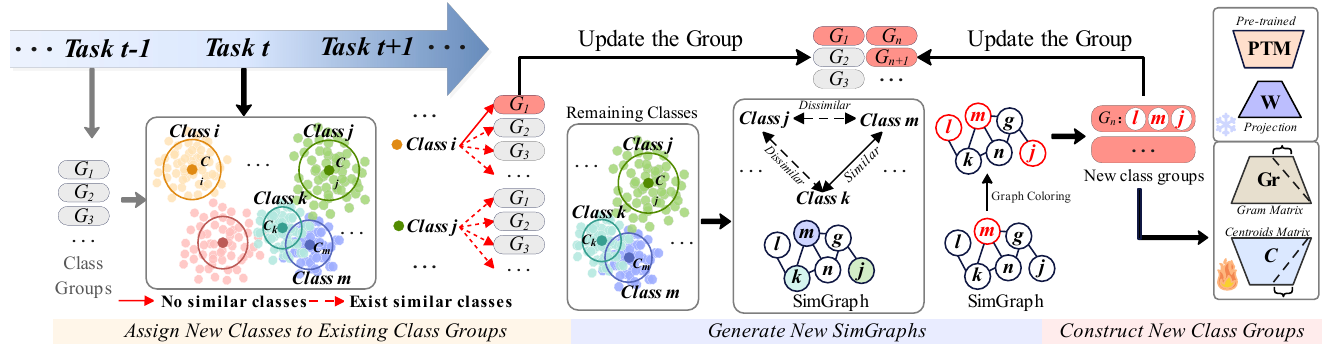}
    \caption{Illustration of The Overall Framework. [best view in color]}
    \label{fig: framework}
\end{figure*}
\textbf{Overview.} \autoref{fig: framework} provides an overview of our proposed method. 
Using task \( t \) as an example, we begin by projecting all training samples into an embedding space utilizing a pre-trained backbone. In this space, we compute the centroids for each class. Next, we evaluate whether a new centroid \( \mathbf{c}_i \) should be integrated into an existing class group \( G_j \).
If \( \mathbf{c}_i \) is dissimilar to all classes within \( G_j \), it is added to the group. If it is similar to any class in an existing group, it remains unassigned.
For unassigned centroids, we construct new similarity graphs (SimGraphs) based on their pairwise similarities. We then apply graph coloring theory to these SimGraphs, forming new class groups by clustering dissimilar categories together.
Finally, we update the NCM-based classifier with all class groups, facilitating efficient model updates with minimal computational overhead.

\subsection{Class Grouping Based on Similarity}

\autoref{Corollary: cor} provides guidance for constructing a sequence of dissimilar tasks. A key idea is to dynamically assign each new class to a group during CIL, ensuring that the similarity between the new class and other classes within the group is minimized. This approach helps maintain the robustness of each group's incremental learning process to the order of tasks. For each group, a separate adapter can be trained, and the results from different adapters can be merged during prediction to enhance the model's overall performance. 

In a given CIL task sequence, we organize the classes into several groups. The group list is denoted as \( G = [G_1, \dots, G_k] \), where each \( G_i \) represents a distinct group of classes. For a specified task \( t \) and each class \( C \in CLS^t \), our objective is to assign class \( C \) to an optimal group \( G^* \), ensuring that the new class is dissimilar to all existing classes in that group.

To achieve this objective, we first define the similarity between classes.
The similarity between any two classes, \( CLS_i \) and \( CLS_j \), is determined using an adaptive similarity threshold \( \eta_{i,j} \).
This threshold is computed based on the mean distance between the training samples of each class and their respective centroids in a learned embedding space, as shown below:

\begin{align}
    \eta_{i,j} = \max [
    & \frac{\sum_{k = 1}^{|X^t|} \mathbb{I}(y^t_k = i) \, d(h(x_k^t), \mathbf{c_i}) }{\sum_{k = 1}^{|X^t|} \mathbb{I}(y^t_k = i)}, \nonumber \\
    & \frac{\sum_{k = 1}^{|X^t|} \mathbb{I}(y^t_k = j) \, d(h(x_k^t), \mathbf{c_j}) }{\sum_{k = 1}^{|X^t|} \mathbb{I}(y^t_k = j)} 
    ],
\end{align}
where \( \mathbf{x}^{(t)} \) denotes the t-th task instance, \( h(\cdot) \) is the feature extraction function defined in Equation \autoref{eq: feature}, \( d: \mathcal{X} \times \mathcal{X} \to \mathbb{R}^+ \) specifies the distance metric space, \( \mathbb{I}(\cdot) \) represents the characteristic function, and the class centroid \( \mathbf{c}_i \in \mathbb{R}^d \) is computed as \( \mathbf{c}_i = \frac{1}{|C_i|} \sum_{x_j \in C_i} \mathbf{x}_j \).

Building upon this framework, we define the condition under which two classes, \( CLS_i \) and \( CLS_j \), are considered dissimilar. Specifically, they are deemed dissimilar if the following condition holds:

\begin{equation}
    d(\mathbf{c_i}, \mathbf{c_j}) > \eta_{i,j}.
    \label{eq: sim}
\end{equation}

Thus, class \( C \) is assigned to group \( G^* \) only if it is dissimilar to all classes within \( G^* \), and \( G^* \) is the choice with the lowest average similarity:

\begin{equation}
    G^* = \arg\min_{G} \frac{1}{|G|} \sum_{C' \in G} d(C, C').
\end{equation}

This approach is consistent with the principles outlined in \autoref{Corollary: cor} and ensures the robustness of the model across the entire task sequence.

\subsection{Graph-Driven Class Grouping}

Graph algorithms provide an efficient method for dynamically grouping classes while minimizing intra-group similarity.
In a graph-theoretic framework, classes are represented as nodes, with edge weights quantifying the similarity between them.
The flexibility and analytical power of graph structures allow for dynamic adjustment of class assignments in CIL, facilitating optimal grouping in polynomial time.
This approach significantly enhances the model's robustness and adaptability in incremental learning tasks.

Therefore, we can leverage the similarity between classes to construct a SimGraph, defined as follows:
\begin{definition} \textbf{(SimGraph.)}
A SimGraph can be defined as an undirect graph $SimG = (V, E)$, where $V$ is the set of nodes that represent each class's centroid and $E$ is the set of edges connecting pair of nodes that represent classes that are determined as similar by \autoref{eq: sim}.
\label{SimGraph}
\end{definition}

Then, we aim to partition the vertex set of this graph into subsets, with each subset forming a maximal subgraph with no edges between vertices. This problem can be abstracted as the classic NP-hard combinatorial optimization problem of finding a minimum coloring of the graphs. Let $G^{-1}(\cdot)$ be an assignment of class group identities to each vertex of a graph such that no edge connects two identically labeled vertices (i.e. $G^{-1}(i) \neq G^{-1}(j)$ for all $(i,j) \in E$). We can formulate the minimum coloring for graph $SimG$ as follows:
\begin{equation}
    \mathcal{X}(SimG) = \min | \{ G^{-1}(k) | k \in V\} |, 
    \label{eq: graph}
\end{equation}
where $\mathcal{X}(SimG)$ is called the chromatic number of $SimG$ and $|\cdot|$ denotes the size of the set.

Brooks' theorem \cite{brooks1941colouring} offers an upper bound for the graph coloring problem. To apply this in our context, we must demonstrate that the similarity graphs constructed in CIL meet the conditions required by Brooks' theorem. By doing so, we can establish that the problem is solvable and that the solution converges, ensuring the effectiveness of our grouping and class coloring process in class incremental learning. Without loss of generality, we can make the following assumptions:

\begin{assumption} In the CIL task, class \( C_i \) is randomly sampled without replacement from the set \( \mathcal{U} = \bigcup_{i=1}^{\infty} C_i \), ensuring that \( C_i \neq C_j \) for all \( i \neq j \). The probability that any two classes \( C_i \) and \( C_j \) within the set \( \mathcal{U} \) meet the similarity condition (as described in \autoref{eq: sim}) is denoted by \( p \).
\end{assumption}

In the CIL scenario with \( N \) classes, the probability of forming an odd cycle is given by \(\left( p^2(1-p)^{(N-2)} \right)^N = p^{2N}(1-p)^{N^2-2N}\). Similarly, the probability of forming a complete graph is \(p^{\binom{N}{2}} = p^{\frac{1}{2}N(N-1)}\).
Thus, the probability that the CIL scenario satisfies Brooks' theorem can be expressed as:
\begin{equation}
    P_{\text{Satisfy Brooks}'} = 1 - p^{2N}(1-p)^{N^2-2N} - p^{\frac{1}{2}N(N-1)}.
\end{equation}
\begin{figure}[t]
    \centering
    \includegraphics[width=\linewidth]{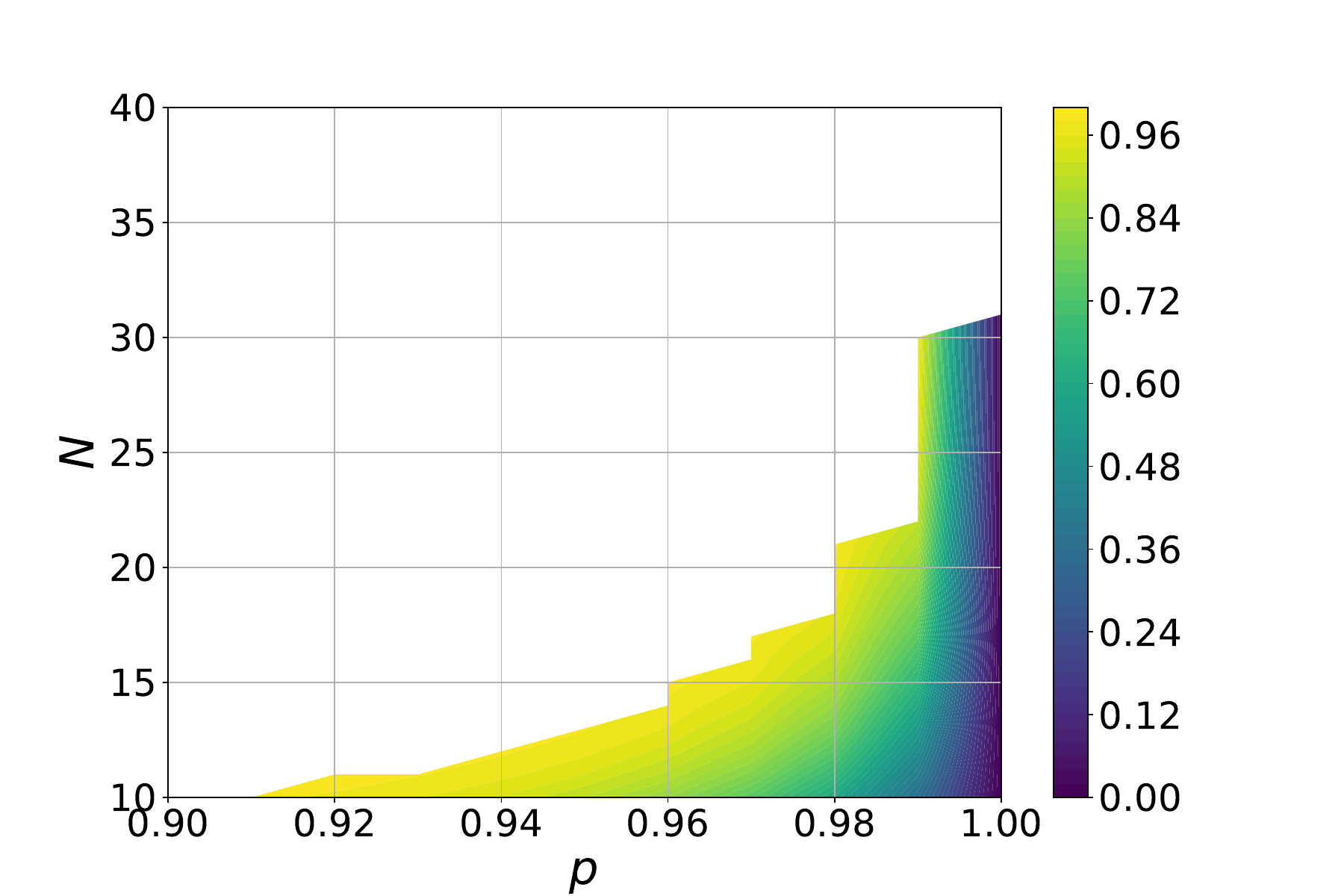}
    \caption{Contour plot delineating the subthreshold region where \( P_{\text{Satisfy Brooks}'} < 0.99 \). The horizontal axis spans \( p \in [0.9, 1.0] \), representing probability values, while the vertical axis specifies sample sizes \( N \in [10, 40] \). In regions not displayed, the corresponding \( P_{\text{Satisfy Brooks}'} \) values exceed 0.99.}
    \label{fig: probability}
\end{figure}

\autoref{fig: probability} illustrates the various values of \( N \) and \( p \) that satisfy Brooks' theorem with a probability of less than 0.99. Our findings indicate that when \( N > 35 \), the CIL scenario adheres to Brooks' theorem. Furthermore, even with fewer classes, as long as \( p \) does not exceed 0.9, the CIL scenario can still ensure that the similarity graph complies with Brooks' theorem at a confidence level of 0.99. We conclude that class grouping based on the similarity graph is convergent and can be solved efficiently in polynomial time.

For \autoref{eq: graph}, while no algorithm exists that can compute \(\mathcal{X}(SimG)\) in polynomial time for all cases, efficient algorithms have been developed that can handle most problems involving small to medium-sized graphs, particularly the similarity graph \(SimG\) discussed here. In practical scenarios, such graphs are typically sparse. Notably, in conjunction with the above analysis, the similarity graph \(SimG\) in the CIL scenario satisfies the non-odd cycle assumption in Brooks' theorem \cite{brooks1941colouring}. For non-complete similarity graphs \(SimG\), we have \(\mathcal{X}(SimG) \le \Delta(SimG)\), where \(\Delta(SimG)\) represents the maximum vertex degree in \(SimG\).

Therefore, we can apply a simple yet effective greedy method, the Welsh-Powell graph coloring algorithm \cite{welsh1967upper}. This algorithm first sorts all nodes in the graph in descending order based on their degree and then assigns a color to each node, prioritizing those with higher degrees. During the coloring process, the algorithm selects the minimum available color for each node that differs from its neighbors, creating new color classes when necessary. The time complexity of this algorithm is \( O(|V|^2) \), primarily due to the color conflict check between each node and its neighbors. In theory, the maximum number of groupings produced by this algorithm is \( \max_{i = 1}^n \min\{ \deg(v_i') + 1, i \} \), with an error margin of no more than 1, where \( V' \) is the sequence of nodes sorted by degree, derived from \( V \).

\subsection{Overall Process}

\noindent \textbf{Training Pipeline.}
Building upon the theoretical foundations in Section 3.1, we now formalize the complete training procedure. Our framework leverages a frozen pre-trained feature extractor $\phi(\cdot)$, augmented with trainable random projections $W \in \mathbb{R}^{L \times M}$ where $M \gg L$, to enhance representation capacity. For each input $x_i^t$ from class group $s$, we compute its expanded feature:
\begin{equation}
\label{eq: feature}
h(x_i^t) = g(\phi(x_i^t) W) \in \mathbb{R}^M,
\end{equation}
where $g(\cdot)$ denotes the nonlinear activation.

The core learning paradigm reframes classification as regularized least-squares regression. Let $H_s^t \in \mathbb{R}^{N_s^t \times M}$ be the feature matrix and $Y_s^t \in \mathbb{R}^{N_s^t \times L_s^t}$ denote the one-hot label matrix for class group $s$. We optimize the projection matrix $\Theta_s^t \in \mathbb{R}^{M \times L_s^t}$ through:
\begin{equation}
\label{eq: loss}
\min_{\Theta} \|Y_s^t - H_s^t \Theta_s^t\|_F^2 + \lambda \|\Theta_s^t\|_F^2,
\end{equation}
where $\lambda$ controls regularization strength. The closed-form solution is:
\begin{equation}
\label{eq: analytic}
\Theta_s^t = ( {H_s^t}^\top H_s^t + \lambda I )^{-1} {H_s^t}^\top Y_s^t.
\end{equation}

For incremental updates, we maintain two key components: the Gram matrix $Gram_s^t$ capturing feature correlations, and the prototype matrix $C_s^t$ encoding class centroids. When new task $t$ arrives with $N_s^t$ samples:
\begin{equation}
    \label{eq: gram}
    Gram_{s}^t = Gram_{s}^{t-1} + \sum_{n = 1}^{N_{s}^t} h(x^t_i)^\top h(x^t_i),
\end{equation}
\begin{equation}
    \label{eq: pro}
    C_{s}^t = \begin{bmatrix}C_{s}^{t-1} & \underbrace{\mathbf{0}_M \ \mathbf{0}_M \ \ldots \ \mathbf{0}_M}_{(L_{s}^t - L_s^{t-1})\text{ times}} \end{bmatrix} + \sum_{n = 1}^{N_s^t} h(x^t_i)^\top y(x^t_i).
\end{equation}

The regularization parameter $\lambda$ is adaptively selected from a candidate pool $\Lambda$ through cross-validation on a held-out calibration set, minimizing the empirical risk:
\begin{equation}
\lambda^* = \arg \min_{\lambda \in \Lambda} \|Y_{\text{val}} - H_{\text{val}} (Gram_{\text{val}} + \lambda I)^{-1} C_{\text{val}} \|_F^2.
\end{equation}

Additionally, group descriptors are constructed through prototype similarity analysis. For each training instance $(x, y) \in \mathcal{D}^t$, we generate meta-features dataset as:
\begin{equation}
\mathcal{D}_g = \left\{ \left( \rho(x),\ G^{-1}(y) \right) \right\}_{(x,y)\in \mathcal{D}^t},
\end{equation}
where $\rho(x) = \big[ d(h(x),\mathbf{c}_1), \ldots, d(h(x),\mathbf{c}_k) \big]^\top$ denotes the concatenated distance vector measuring similarity between the sample embedding and prototype vectors.

\noindent \textbf{Inference Pipeline.}
Given test sample $x^*$, its group identification can be learned via $\hat{g} = \mathcal{M}_g(\rho(x^*))$, where $\mathcal{M}_g$ is the class group predict model trained with $ \mathcal{D}_g$.
Then, the prediction will be performed within the selected group via $\hat{y} = \underset{c \in \mathcal{C}_{\hat{g}}}{\arg\max}\ ( g(\phi(x^*) W) (Gram_{\hat{g}} + \lambda I)^{-1} C_{\hat{g}}[:,c]$.

\section{Experiment}

\begin{table*}[ht] \footnotesize
    \caption{Results (\%) of CL Methods on Both Fine-grained Datasets and General Vision Dataset (Split into 10 Tasks). Among Them, The Best Results Are Bolded for Emphasis, While The Second-best Results Are Underlined.}
    \label{tab:mainresult}
    \centering
    \begin{tabular*}{1\linewidth}{*{10}{c}}
        \toprule
        \multirow{2}{*}{Method} & \multicolumn{2}{c}{CIFAR100} & \multicolumn{2}{c}{CUB200} & \multicolumn{2}{c}{Dog} & \multicolumn{2}{c}{OB} \\
        \cmidrule(lr){2-3} \cmidrule(lr){4-5} \cmidrule(lr){6-7} \cmidrule(lr){8-9} 
        & $A_N$ ($\uparrow$) & $F_N$ ($\downarrow$) & $A_N$ ($\uparrow$) & $F_N$ ($\downarrow$)& $A_N$ ($\uparrow$) & $F_N$ ($\downarrow$)& $A_N$ ($\uparrow$) & $F_N$ ($\downarrow$) \\
        \midrule
        Finetune & 67.86 $\pm$ 0.46 & 31.25 $\pm$ 1.76 & 48.91 $\pm$ 0.49 & 47.68 $\pm$ 3.48 & 45.64 $\pm$ 4.40 & 46.19 $\pm$ 3.68 & 61.73 $\pm$ 5.70 & 31.84 $\pm$ 4.87 \\ 
        
        L2P & 83.45 $\pm$ 0.25 & 8.50 $\pm$ 0.43 & 66.57 $\pm$ 1.37 & 12.00 $\pm$ 1.41 & 65.09 $\pm$ 3.74 & 9.44 $\pm$ 2.93 & 70.10 $\pm$ 8.86 & 13.70 $\pm$ 3.39 \\ 
        
        DualPrompt & 82.74 $\pm$ 0.85 & 7.19 $\pm$ 1.18 & 68.21 $\pm$ 1.55 & 12.08 $\pm$ 4.65 & 70.90 $\pm$ 0.57 & 10.68 $\pm$ 1.35 & 69.73 $\pm$ 8.26 & 11.91 $\pm$ 2.28 \\
        
        CODA-Prompt & 86.86 $\pm$ 4.26 & 6.04 $\pm$ 0.71 & 73.91 $\pm$ 1.45 & 7.84 $\pm$ 0.10 & 74.09 $\pm$ 0.69  & 10.05 $\pm$ 0.07 & 77.59 $\pm$ 8.41  & 9.05 $\pm$ 1.89 \\
        
        SimpleCIL & 77.63 $\pm$ 2.01 & 7.21 $\pm$ 0.48 & 83.32 $\pm$ 2.03 & 5.89 $\pm$ 1.26 & 83.23 $\pm$ 0.72 & 5.86 $\pm$ 0.43 & 72.15 $\pm$ 0.02 & 8.31 $\pm$ 0.31 \\
        
        ADAM & 84.15 $\pm$ 0.77 & 5.03 $\pm$ 0.15 & 83.89 $\pm$ 2.07 & 5.86 $\pm$ 1.12 & 83.98 $\pm$ 0.34 & 6.08 $\pm$ 0.54 & 72.84 $\pm$ 0.29 & 8.23 $\pm$ 0.16 \\
        
        EASE & 86.95 $\pm$ 0.41 & 6.74 $\pm$ 0.43 & 85.78 $\pm$ 4.22 & 6.27 $\pm$ 2.98 & 81.03 $\pm$ 1.38 & 8.18 $\pm$ 0.50 & 74.56 $\pm$ 4.32 & 8.60 $\pm$ 1.40 \\
        
        RanPAC & \underline{90.50 $\pm$ 0.34} & \underline{3.49 $\pm$ 0.16} & \underline{89.23 $\pm$ 0.36} & \underline{3.91 $\pm$ 0.45} & \underline{85.37 $\pm$ 0.41} & \underline{5.78 $\pm$ 0.42} & \underline{78.86 $\pm$ 0.16} & \underline{7.62 $\pm$ 0.18} \\ 
        
        GDDSG (Ours) & \textbf{94.00 $\pm$ 0.03} & \textbf{0.78 $\pm$ 0.09} & \textbf{91.64 $\pm$ 1.86} &\textbf{ 1.92 $\pm$ 1.29} & \textbf{92.64 $\pm$ 0.48} & \textbf{1.42 $\pm$ 0.08} & \textbf{87.33 $\pm$ 0.32} & \textbf{0.96 $\pm$ 0.16} \\
        \bottomrule
    \end{tabular*}
\end{table*}

\subsection{Experimental Setup}\label{5.1}

\textbf{Datasets.} Since most pre-trained models are currently trained on ImageNet-21K \cite{deng2009imagenet}, we aim to assess the model’s performance on entirely new data. To demonstrate the robustness of our model to task similarity, we conduct experiments using several datasets, including CIFAR100 \cite{krizhevsky2009learning}, CUB200 \cite{wah2011caltech}, Stanford Dogs \cite{KhoslaYaoJayadevaprakashFeiFei_FGVC2011}, and OmniBenchmark (OB) \cite{zhang2022benchmarking}. These datasets are divided into multiple, equally sized tasks, and various class orders are tested to evaluate the model’s performance across different orders.

\noindent \textbf{Baseline.} For fairness, we only compare against CL methods that have utilized pre-trained models in recent years. We compare GDDSG with the following six latest and effective CL methods with the PILOT toolbox \cite{sun2023pilot}: L2P \cite{wang2022learning}, Dualprompt \cite{wang2022dualprompt}, CODA-Prompt \cite{smith2023coda}, SimpleCIL \cite{zhou2023revisiting}, ADAM \cite{zhou2023learning}, EASE \cite{zhou2024expandable}, RanPAC \cite{mcdonnell2024ranpac}.

\noindent \textbf{Implementations.} Our code, implemented in PyTorch, has been open-sourced for accessibility. All experiments were conducted on a single Nvidia RTX 3090 GPU, using three random seeds, 1993, 2024, and 4202, to compute the average for a more robust model evaluation. We use a ViT-B/16 model, which is self-supervised and pre-trained on ImageNet-21K.
$\mathcal{M}_g$ is a soft voting model consisting of RandomForest \cite{breiman2001random}, KNN \cite{peterson2009k}, and LightGBM \cite{ke2017lightgbm}.
%Detailed dataset descriptions and experimental implementations are provided in the Supplementary Material.

\noindent \textbf{Metrics.} We employ average final accuracy $A_N$ and average forgetting rate $F_N$ as metrics \cite{wang2022learning}. $A_N$ is the average final accuracy concerning all past classes over $N$ tasks. $F_N$ measures the performance drop across $N$ tasks, offering valuable information about plasticity and stability during CL. 
Following the protocol in \cite{lian2022scaling}, we use the Order-normalized Performance Disparity (OPD) metric to assess the robustness of the class order. OPD is calculated as the performance difference of task \( t \) across \( R \) random class orders, defined as:
\begin{equation}
    \text{OPD}_t = \max\{ \bar{A}_t^1, \ldots, \bar{A}_t^R \} - \min\{ \bar{A}_t^1, \ldots, \bar{A}_t^R \}.
\end{equation}

To characterize both extreme scenarios and overall performance, we introduce two consolidated metrics: Maximum OPD (MOPD) defined as \( \text{MOPD} = \max\left\{\text{OPD}_1, \ldots, \text{OPD}_T\right\} \), and Average OPD (AOPD) calculated by \( \text{AOPD} = \frac{1}{T} \sum_{t=1}^{T} \text{OPD}_t \).
\iffalse
The Maximum OPD (MOPD) and Average OPD (AOPD) are further defined as:
\begin{equation}
    \text{MOPD} = \max\{\text{OPD}_1, \ldots, \text{OPD}_T\},
\end{equation}
\begin{equation}
    \text{AOPD} = \frac{1}{T} \sum_{t=1}^{T} \text{OPD}_t.
\end{equation}
\fi
\subsection{Experimental Results}\label{5.2}

\textit{\textbf{Main Results.}}  \autoref{tab:mainresult} highlights the strong performance of our proposed GDDSG method in terms of accuracy and resistance to forgetting. The results demonstrate that GDDSG consistently outperforms other techniques, achieving state-of-the-art (SOTA) performance. Notably, GDDSG shows marked improvements in both accuracy and forgetting rate. Compared to the previous SOTA method, RanPAC, our approach achieves significantly higher accuracy while maintaining a low forgetting rate of around 1\%, underscoring GDDSG's superior effectiveness in the CIL environment.

\noindent \textit{\textbf{Ablation analysis.}} Our method's two components, SimGraphs and Class Groups, operate as a unified whole. Only after generating the SimGraphs can construct the Class Groups. Therefore, we can only conduct ablation experiments on either individual Class Groups or the SimGraphs and Class Groups combination as a whole, with results shown in \autoref{tab: ablation}. The results demonstrate a significant decrease in model performance after conducting the ablation, validating the effectiveness of SimGraphs and Class Groups.

\begin{table}[t]
    \centering
    \caption{Ablation Experiment.}
    \label{tab: ablation}\small
    \begin{tabular}{cccccc}
        \toprule
        \textbf{$A_N$} & CIFAR100 & CUB200 & Dog & OB \\
        \midrule
        w/o Class Groups & 74.32 & 72.86 & 69.49 & 66.85\\
        \makecell[c]{w/o SimGraphs and\\Class Groups}  & 89.96 & 87.32 & 83.12 & 74.21\\
        GDDSG & \textbf{93.99} & \textbf{92.95} & \textbf{92.30} & \textbf{87.56}\\
        \midrule
        \textbf{$F_N$} & CIFAR100 & CUB200 & Dog & OB \\
        \midrule
        w/o Class Groups & 12.42 & 16.70 & 14.13 & 20.95\\
        \makecell[c]{w/o SimGraphs and\\Class Groups} & 4.12 & 3.78 & 5.99 & 9.46\\
        GDDSG & \textbf{0.72} & \textbf{1.01} & \textbf{1.48} & \textbf{1.07} \\
        \bottomrule
    \end{tabular}
\end{table}

\begin{figure}
    \centering
    \includegraphics[width=\linewidth]{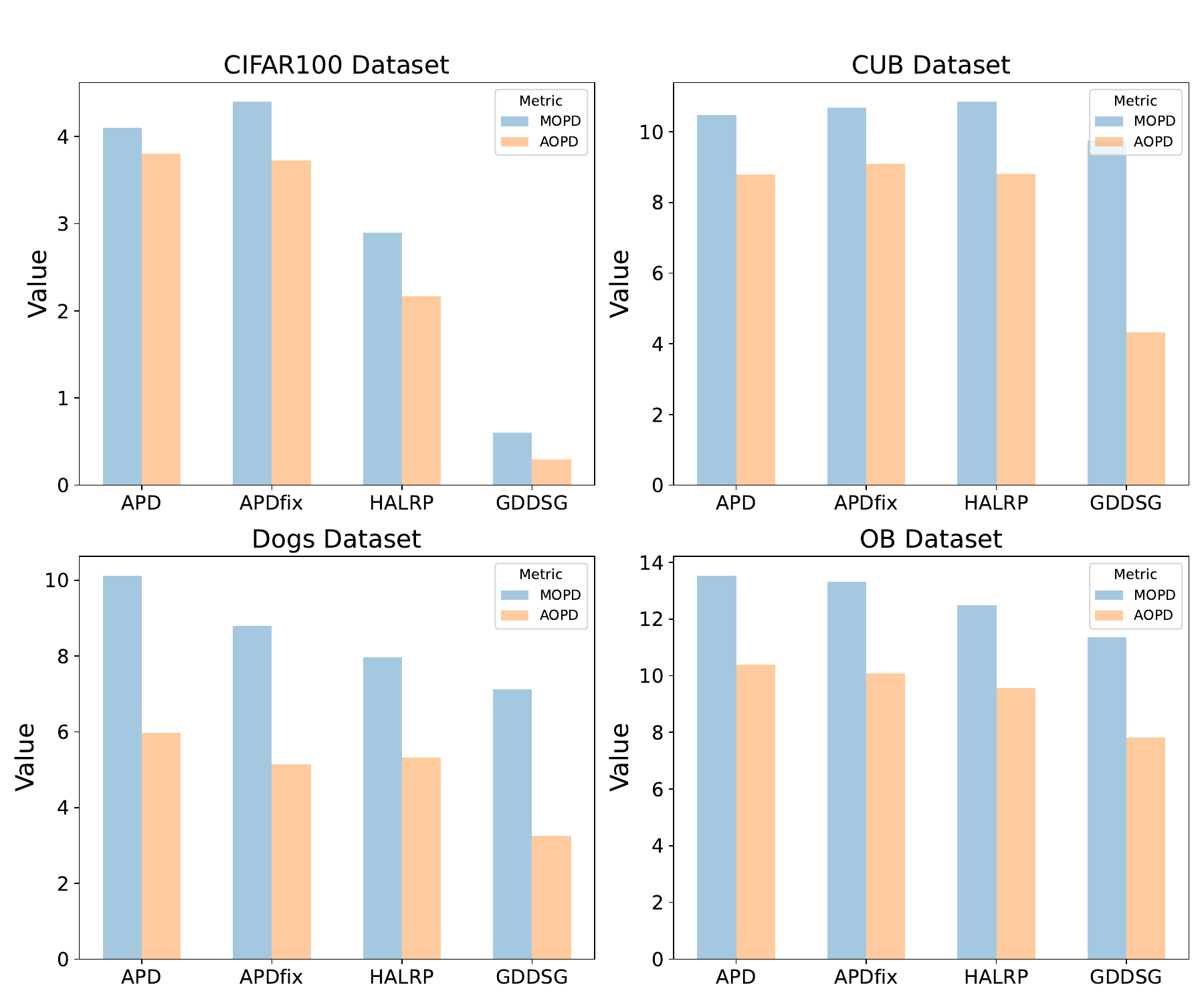}
    \caption{Robustness of Different Methods to Class Order: MOPD (Blue) and AOPD (Orange) Indicators, with GDDSG Performing Well Across Four Datasets}
    \label{fig: Robustness}
\end{figure}

\noindent \textit{\textbf{Robustness to Class Order.}} We conducted comparative experiments on existing order-robust CIL methods, including APD \cite{lian2022scaling}, APDfix \cite{lian2022scaling}, and HALRP \cite{li2024hessian}, using 10 different class orders across four datasets and calculating their MOPD and AOPD metrics. The experimental results, presented in \autoref{fig: Robustness}, show that our proposed GDDSG method demonstrates excellent robustness to category order. MOPD decreased across all datasets, with AOPD showing a significant reduction, underscoring the practical effectiveness of our approach.

\begin{figure}[ht]
  \centering
  \includegraphics[width=0.47\textwidth]{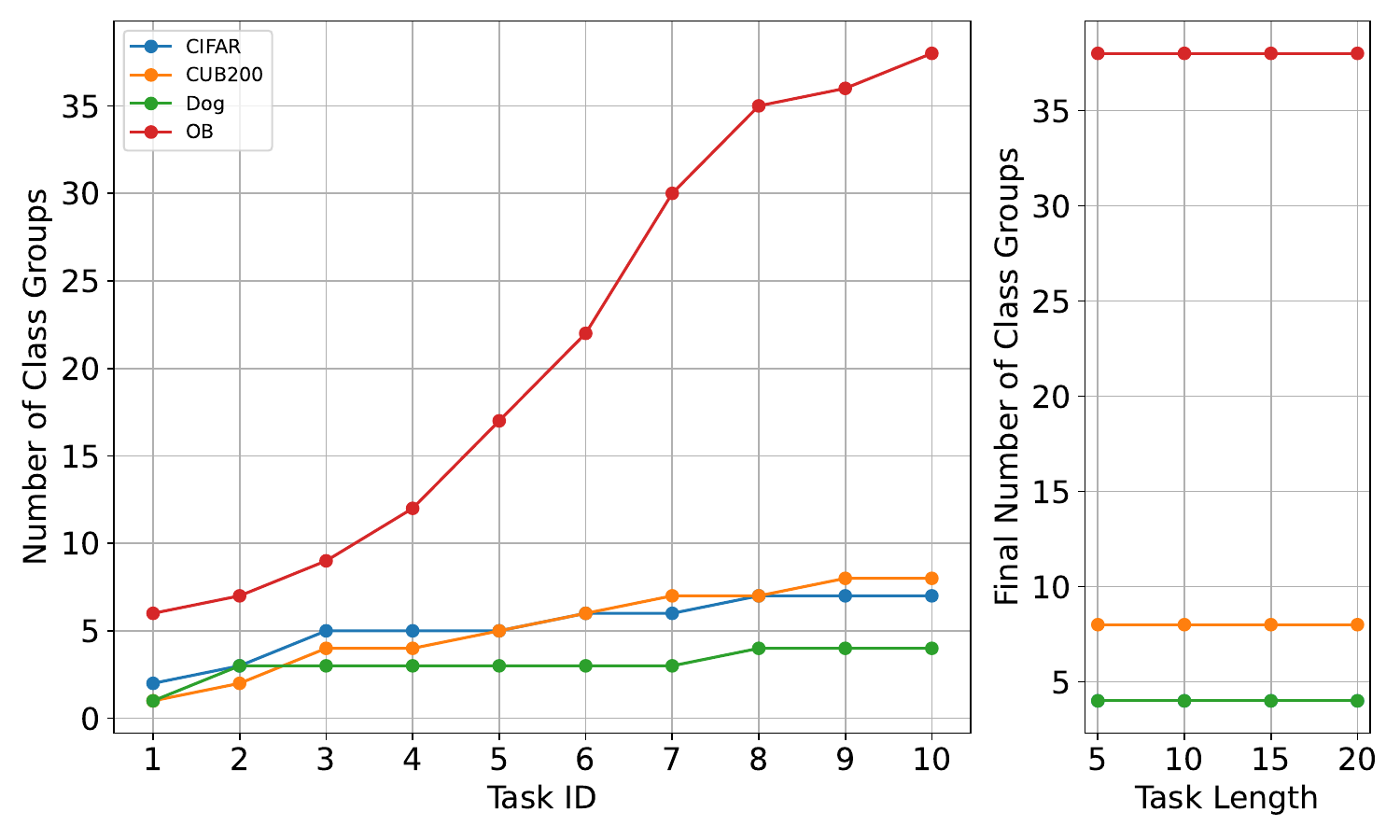}
  \caption{Analysis of Class Group Counts: The Left Figure Shows Changes in Class Group Counts as the Number of Tasks Increases, and the Right Figure Shows Changes as Task Length Varies.}
  \label{fig: ex1}
\end{figure}
\noindent \textit{\textbf{Analysis of Class Group Counts.}} \autoref{fig: ex1} illustrates how the number of class groups changes as tasks arrive and task lengths vary during the execution of the GDDSG algorithm. We observe that for relatively homogeneous datasets, such as CIFAR-100 and CUB-200, the optimal number of class groups generated remains relatively low and tends to stabilize midway through the task sequence. In contrast, datasets with broader domains and more categories, such as OB, result in a higher number of optimal class groups. Despite this increase, GDDSG maintains accurate matching, demonstrating its strong generalization capability.

Additionally, we simulate varying frequencies of intra-task class conflicts by altering the number of categories within a single task, which leads to differences in both intra-task and inter-task similarities. The results indicate that the optimal number of class groups determined by the GDDSG algorithm consistently converges to a stable value. This demonstrates that, for a specific dataset and pre-trained model, the optimal number of class groups is determined solely by the dataset itself, independent of factors such as task length and order in the CIL environment. This stability arises because the GDDSG algorithm primarily relies on data similarity, while disregarding task-specific information in real-world scenarios. The robustness of this approach is theoretically supported by Brooks' theorem \cite{brooks1941colouring} and the Welsh–Powell algorithm \cite{welsh1967upper}. In conclusion, the GDDSG algorithm exhibits strong robustness to variations in task similarity, length, and order, making it highly valuable for a wide range of applications.

\begin{table}[t]
    \caption{The Class Group Number And Final Results of GDDSG Under Different Pre-trained Backbones.}
    \label{tab:backbone}
    \centering\footnotesize
    \begin{tabular}{>{\centering}m{1.2cm} >{\centering}m{1.2cm} >{\centering}m{1.3cm} >{\centering}m{1.3cm} >{\centering\arraybackslash}m{1.3cm}}
        \toprule
        \multirow{2}{*}{\centering Dataset} & \multirow{2}{*}{\centering Metric} & \multicolumn{3}{c}{Backbone} \\
        \cmidrule(lr){3-5}
        & & ViT B/16 & ResNet-50 & ResNet-18 \\
        \midrule
        \multirow{3}{*}{\centering CIFAR100} & $ \mathcal{X}$ ($\downarrow$) & \textbf{7}  & 39 & 34 \\
        & $A_N$ ($\uparrow$) & \textbf{93.99}  & 83.25 & 80.97 \\
        & $F_N$ ($\downarrow$) & 0.72  & 0.73 & \textbf{0.63} \\
        \midrule
        \multirow{3}{*}{\centering CUB200} & $ \mathcal{X}$ ($\downarrow$) & \textbf{8}  & 34 & 42 \\
        & $A_N$ ($\uparrow$) & \textbf{92.95}  & 70.83 & 51.44 \\
        & $F_N$ ($\downarrow$) & \textbf{1.01}  & 2.43 & 18.32 \\
        \midrule
        \multirow{3}{*}{\centering Dog} & $ \mathcal{X}$ ($\downarrow$) & 4  & \textbf{3} & 18 \\
        & $A_N$ ($\uparrow$) & \textbf{92.30}  & 82.65 & 70.56 \\
        & $F_N$ ($\downarrow$) & \textbf{1.48}  & 1.74 & 5.34 \\
        \midrule
        \multirow{3}{*}{\centering OB} & $ \mathcal{X}$ ($\downarrow$) & 38  & \textbf{36} & 44 \\
        & $A_N$ ($\uparrow$) & \textbf{87.56}  & 77.24 & 62.35 \\
        & $F_N$ ($\downarrow$) & \textbf{1.07}  & 1.17 & 2.14 \\
        \bottomrule
    \end{tabular}
\end{table}

\noindent \textit{\textbf{Detailed Analysis of Backbone.}} \autoref{tab:backbone} presents the optimal number of class groups that GDDSG generates under various pre-trained backbone networks, along with the corresponding \( A_N \) and \( F_N \) values. The results indicate that smaller backbones, such as ResNet18 and ResNet50, yield reduced accuracy and higher forgetting rates compared to using ViT as the backbone. Nonetheless, performance with ResNet50 remains highly competitive, achieving accuracy comparable to L2P while maintaining a relatively low forgetting rate. This further highlights the robustness of GDDSG, as it can reach performance levels similar to those of richer, more powerful backbones, even when using networks with fewer parameters and lower representational capacity.

Another noteworthy observation is that, for the same dataset, the number of class groups varies depending on the backbone network. Specifically, when using ResNet18 or ResNet50, the number of class groups for CIFAR-100 and CUB-200 increases significantly, whereas it decreases for Stanford Dogs and OB. This suggests that, for certain datasets, backbones with lower representational power may erroneously classify some originally dissimilar classes as similar, leading to an increase in class groups. In contrast, other datasets remain unaffected by the choice of backbone. This highlights that the impact of the backbone on the GDDSG algorithm is highly dataset-dependent.

\begin{figure}[t]
  \centering
  \includegraphics[width=0.47\textwidth]{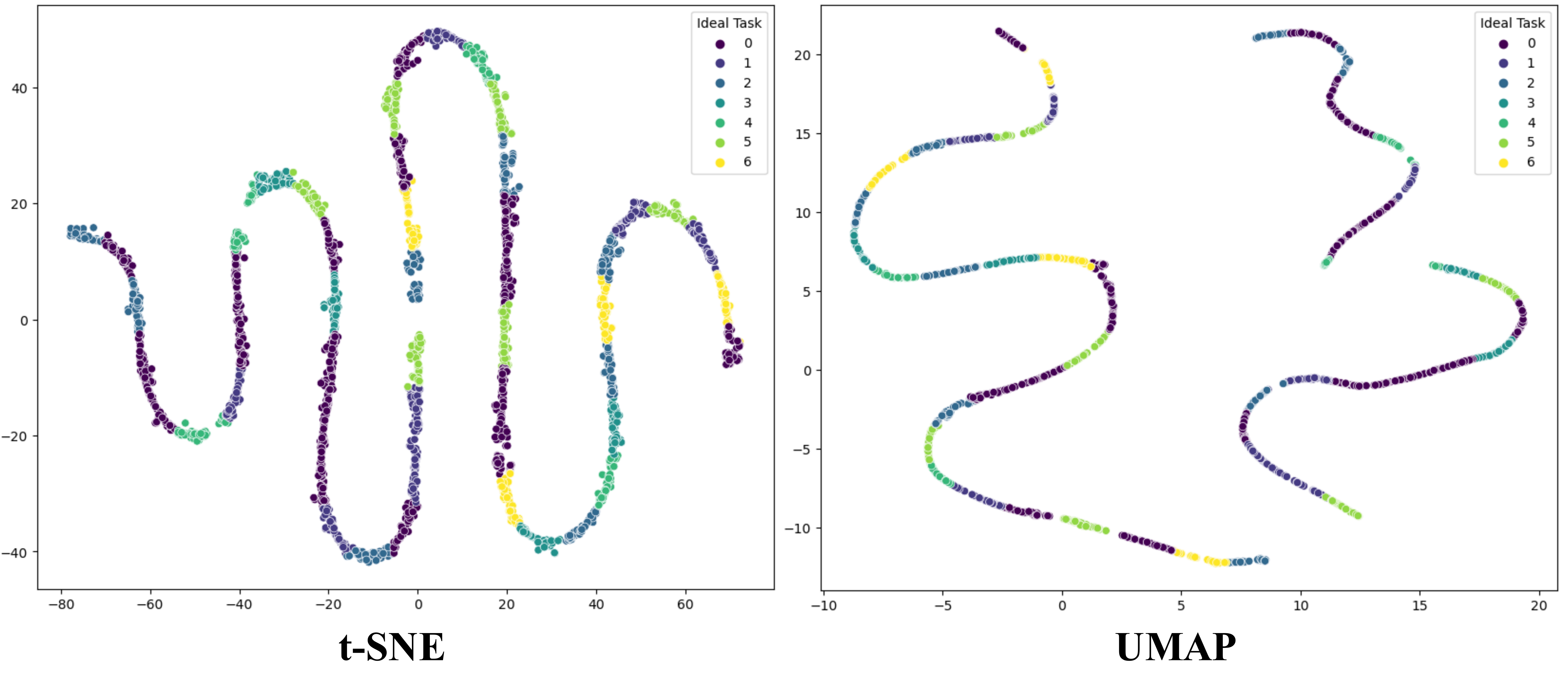}
  \caption{Visualization of The Distance Feature via T-SNE and UMAP After Training on All Tasks (Split CIFAR100 Dataset).}
  \label{fig:ex2}
\end{figure}

\noindent \textit{\textbf{Detailed Analysis of Class Group Identification.}} To better understand the mechanism of class group identification, we visualized $\mathcal{D}_g$ using t-SNE \cite{van2008visualizing} and UMAP \cite{McInnes2018UMAPUM} on the Split CIFAR100 dataset. \autoref{fig:ex2} presents the visualized results. The observations reveal that the distance features essentially conform to piecewise functions in high dimensions, exhibiting strong linear separability and powerful representation capabilities. Consequently, a class group identification matching model can be effectively fitted using some classical machine learning models, enabling fairly accurate predictions. However, it is crucial to emphasize the importance of this step for GDDSG. Under the partition of the GDDSG algorithm, the accuracy of a single class group can reach nearly 100\%. Therefore, the precision of class group matching directly determines the overall model's accuracy.

\section{Conclusion \& Limitation}
\iffalse
In this study, we aim to design an order-robust CIL model capable of addressing two critical challenges: class order sensitivity and intra-task conflicts. Building on existing theories, we find that as class similarity decreases, the model's sensitivity to class order also lessens, which effectively mitigates knowledge conflicts both across tasks and within individual tasks. 
To enhance the model's robustness across varying class orders, we propose a dynamic grouping method based on similarity graphs, termed GDDSG.
The proposed approach maintains the centroids of learned classes and group classes based on dynamic similarity. In GDDSG, we introduce a novel approach to structuring class groups within class-incremental learning. Our GDDSG can continually update existing groups or form new ones, training distinct models for each group. During inference, predictions are derived through an ensemble of outputs from multiple models, thereby enhancing overall accuracy and robustness in CIL.
\fi
This study addresses the critical challenge of class order sensitivity in Class Incremental Learning (CIL), where model performance significantly degrades under varying class arrival sequences. By introducing GDDSG, a graph-driven framework that dynamically partitions classes into similarity-constrained groups and coordinates isolated sub-models with joint prediction, we theoretically and empirically mitigate the impact of class sequence variations. Experiments validate that our method not only reduces sensitivity to class order but also achieves state-of-the-art accuracy and anti-forgetting performance. This work provides a benchmark for developing robust CIL methods with dynamic data streams.

Inevitably, our method has certain limitations. First, GDDSG currently relies on NCM classifiers. In future work, we aim to explore order-robust CIL approaches with Softmax strategies. 
Also, while the memory overhead remains small, it could be further streamlined for efficiency, and we intend to address this limitation with future studies.

\section*{Acknowledgement}
This work was supported by the National Natural Science Foundation of China (No. 62476228, No. 72242106), the Sichuan Science and Technology Program (No. 2024ZYD0180), and the Major Basic Research Project of Shandong Provincial Natural Science Foundation (No. ZR2024ZD03).
{
    \small
    \bibliographystyle{ieeenat_fullname}
    \bibliography{main}
}

\newpage
\appendix
\section*{Appendix}

\setcounter{table}{0} 
\setcounter{figure}{0}
\setcounter{equation}{0}
\setcounter{algocf}{0}  

%定义编号格式，在数字序号前加字符“A."
\renewcommand\thetable{A.\arabic{table}}
\renewcommand\thefigure{A.\arabic{figure}}
\renewcommand\theequation{A.\arabic{equation}}
\renewcommand\thealgocf{A.\arabic{algocf}} 

\renewcommand*{\theHtable}{\thetable}
\renewcommand*{\theHfigure}{\thefigure}
\renewcommand*{\theHequation}{\theequation}
\renewcommand*{\theHsection}{\thesection}
\renewcommand*{\theHalgocf}{\thealgocf}

\section{Notation}\label{notation}

In \autoref{tab: notations}, we introduce the notations throughout this paper.
\begin{table}[ht]
 \begin{center}\footnotesize
  \begin{tabular}{c c}
    \hline
    \textbf{Notation} & \textbf{Explanation} \\ 
    \hline
    $t$ & The task $t$ \\
    $\mathcal{D}^t$ & The training set of task $t$\\
    $X^t$ & The set of training samples of task $t$\\
    $Y^t$ & The set of training samples labels of task $t$\\
    $CLS^t$ &  The class set of task $t$\\
    $|CLS^t|$ &  The class account of task $t$\\
    $\phi(\cdot)$ & The CL model's feature extractors\\
    $f(\cdot)$ & The CL model's classifier\\
    $c_i$ & The centroid for class \\
    $G$ & The class group\\
    $\eta_{i,j}$ & The adaptive similarity threshold for $CLS_i$ and $CLS_j$\\
    $d(\cdot,\cdot)$ & The distance function \\
    $\mathbb I(\cdot)$ & The indicator function \\
    $SimG$ & The graph generated base on \autoref{eq: sim} \\
    $V$ & The node set of $SimG$\\
    $E$ & The edge set of $SimG$\\
    $B(\cdot)$ & The corresponding class group to each vertex of a $SimG$\\
    $\mathcal{X}(SimG)$ & The chromatic number of $SimG$\\
    $\Delta(SimG)$ & The maximum degree of vertices in $SimG$\\
    $W$ & The random projection matrix\\
    $L$ & The dimension size of $\phi$\\
    $M$ & The dimension size after random projection\\
    $h(x_i^t)$ & The feature vector of the sample $x_i^t$\\
    $y(x_i^t)$ & The one-hot label embedding of the sample $x_i^t$\\
    $g(\cdot)$ & The nonlinear function\\
    $H_s$ & The concatenation of feature vector of class group $s$\\
    $Gr_s$ & The Gram matrix of class group $s$\\
    $L_s^t$ & The number of class for class group $s$ until task $t$\\
    $N_s^t$ & The number of samples for class group $s$ in task $t$\\
    $PIT(\cdot)$ & The function to predict the class group\\
    $p_i$ &  Probability outputted by Softmax\\
    $s_i$ &  Scores outputted by NCM \\
    \hline
  \end{tabular}
  \caption{Notations and explanations.} 
  \label{tab: notations}
 \end{center}
\vspace{-0.5cm}
\end{table}

\section{Datesets, Implementations and Additional Experimental Results}

\subsection{Datasets}\label{sec: datasets}

\begin{table}[h]\footnotesize
\begin{tabular}{ccccc}
\hline
Datasets      & Orginal & $N$ & Val samples & Class numbers \\\hline
CIFAR100      &   \cite{krizhevsky2009learning}   & 50000            & 10000        & 100             \\
CUB           &    \cite{wah2011caltech}     & 9430             & 2358         & 200             \\
Stanford Dog  &    \cite{KhoslaYaoJayadevaprakashFeiFei_FGVC2011}     & 12000            & 8580         & 120             \\
OmniBenchmark &    \cite{zhang2022benchmarking}     & 89697            & 5985         & 300   \\  \hline 
\end{tabular}
\caption{Datasets. We list references for the original source of each dataset. In the column headers, $N$ is the total number of training samples, \emph{Class numbers} is the number of classes following training on all tasks, and \# of val samples is the number of validation samples in the standard validation sets.}
\label{tab: datasets}
\end{table}

The four CL datasets we use are summarised in \autoref{tab: datasets}. For CUB and Omnibenchmark we used specific train-validation splits defined and outlined in detail by \cite{zhou2023revisiting}. For the CIFAR100, CUB, and Stanford Dog datasets, which are categorized as fine-grained datasets, their task similarity has a significant impact and is used to measure the knowledge specialization of the model. Omnibenchmark has substantial classes and samples, with diverse sample sources, which can effectively measure the model's knowledge generalization ability.

\subsection{Detail of Metrics}\label{detailofmetrics} 

We employ average final accuracy $A_N$ and average forgetting rate $F_N$ as metrics \cite{wang2022learning}. $A_N$ is the average final accuracy concerning all past classes over $N$ tasks. $F_N$ measures the performance drop across $N$ tasks. We use $Acc_i^t$ to denote the test accuracy of class $i$ after the completion of task $t$ and $Acc_i^{t_0}$ to denote the test accuracy of class $i$ after its first task $t_0$. Accordingly, $A_N$ and $F_N$ can be expressed as:
\begin{equation}
    A_N = \frac{\sum_{i\in{\sum_{t = 1}^T |CLS^t|}} Acc_i^T }{\sum_{t = 1}^T |CLS^t|},
\end{equation}
\begin{equation}
    F_N = \frac{\sum_{i\in{\sum_{t = 1}^T |CLS^t|}} (Acc_i^T - Acc_i^{t_0}) }{\sum_{t = 1}^T |CLS^t|}.
\end{equation}
It is worth mentioning that, unlike the evaluation metrics used in \cite{zhou2023learning,zhou2023revisiting,zhou2024expandable}, our metric ensures that each class has an equal evaluation weight, thereby avoiding increased sensitivity to previous tasks.
Following the protocol in \cite{lian2022scaling}, we use the Order-normalized Performance Disparity (OPD) metric to assess the robustness of the class order. OPD is calculated as the performance difference of task \( t \) across \( R \) random class orders, defined as:
\begin{equation}
    \text{OPD}_t = \max\{ \bar{A}_t^1, \ldots, \bar{A}_t^R \} - \min\{ \bar{A}_t^1, \ldots, \bar{A}_t^R \}.
\end{equation}
The Maximum OPD (MOPD) and Average OPD (AOPD) are further defined as:
\begin{equation}
    \text{MOPD} = \max\{\text{OPD}_1, \ldots, \text{OPD}_T\},
\end{equation}
\begin{equation}
    \text{AOPD} = \frac{1}{T} \sum_{t=1}^{T} \text{OPD}_t.
\end{equation}

\subsection{Training Details}\label{trainingdetails}

We followed the general setting in the continual learning community, i.e., randomly shuffled the session order for each dataset. 
The results presented throughout this paper are the mean results of two random shuffles.
We implement all experiments on one NVIDIA GeForce-RTX-3090 GPU and the Pytorch library. 
Input images are resized to 224 x 224 and normalized to the range of [0,1]. The hyperparameter settings for each baseline are set according to the optimal combination reported in their papers, respectively.
The Adam optimizer trains all Softmax-based models with a batch size of 128 and a learning rate of 0.05. 
The proposed SALF used ViT as the backbone, pre-trained on ImageNet-21k, with frozen parameters except for the classification header. 

Our contribution also includes faithful PyTorch implementations of our method and abundant baselines under the CL setting. 

\subsection{Baselines Description}
We compare our proposed SALF against a wide range of benchmarks on four widely used datasets to thoroughly validate it. SALF outperforms previous works, setting a new state-of-the-art performance.
We provide detailed descriptions of all the baselines:
\begin{itemize}
    \item \emph{Finetune} adjusts classifier weights through cross-entropy loss.
    \item \emph{L2P} selects prompts from the prompt pool using the key query matching strategy.
    \item \emph{DualPrompt} attach prompts to different layers to decompose prompts into universal and expert prompts.
    \item \emph{CODA-Prompt} builds attention-based prompts from the prompt pool.
    \item \emph{SimpleCIL} replace updated model classifier weights with class prototypes.
    \item \emph{ADAM} fine-tuning based on SimpleCIL.
    \item \emph{RanPAC} projects the feature space onto a higher dimensional space to approach a Gaussian distribution and eliminates mutual information between classes through the Gram matrix. 
\end{itemize}

\iffalse
\subsection{Additional Experimental Results}\label{additionalex}

\textbf{Model Selection for Class Group Matching} In the paper, we highlight that the matching of Class Groups determines the model's overall performance. Consequently, we conducted an experimental analysis using several classic, highly interpretable methods. The experiments were performed on the Stanford Dog dataset, where our task was to predict the Class Group of each test sample after partitioning with the SALF algorithm. The results are shown in \autoref{fig: model selection}. We observe that these classic methods achieved excellent performance, indicating that predicting based on the distance matrix between samples and prototypes is feasible. Additionally, due to the outstanding performance of the Random Forest method, we employed it for Class Group matching in our experiments.
\fi
\iffalse
\begin{figure}[t]
    \centering
    \includegraphics[width=0.6\linewidth]{figures/output.png}
    \caption{Accuracy of Class Group Matching (\%). A comparison was made between five classic algorithms: Decision Tree, MLP, LightGBM, KNN, and RandomForest.}
    \label{fig: model selection}
\end{figure}
\fi

\section{Proof of Theorem}\label{proof}

\subsection{Proof of Brooks’ Theorem}

\textbf{Proof}: 

Let $|V(G)| = n$, and we proceed by mathematical induction.

Firstly, when $n \leq 3$, the proposition holds.

Next, assuming the proposition holds for $n-1$, we aim to strengthen it step by step.

Without loss of generality, consider $\Delta(G)$-regular graphs, since non-regular graphs can be seen as obtained by removing some edges from regular graphs, which doesn't affect the conclusion.

For any regular graph $G$ that is neither complete nor an odd cycle, let's take a vertex $v$ and consider the subgraph $H:=G-v$. By the inductive hypothesis, we know that $\mathcal{X} (H) \leq \Delta(H) = \Delta(G)$. Now we only need to prove that inserting $v$ into $H$ does not affect the conclusion.

Let $\Delta:=\Delta(G)$, and suppose $H$ is colored with $\Delta$ colors: $c_1, c_2, ..., c_{\Delta}$. The $\Delta$ neighbors of $v$ are denoted as $v_1, v_2, ..., v_{\Delta}$. Without loss of generality, assume that these neighboring colors of $v$ are pairwise distinct; otherwise, the proposition holds.

Next, let's consider all the vertices colored with either $c_i$ or $c_j$ in $H$, and all edges between them, forming a subgraph $H_{i,j}$. Without loss of generality, assume that any two different vertices $v_i$ and $v_j$ are in the same connected component of $H_{i,j}$. Otherwise, if they were in different connected components, we could exchange the colors of all vertices in one of the connected components, making $v_i$ and $v_j$ have the same color.

We denote the aforementioned connected components as $C_{i,j}$, where $C_{i,j}$ must necessarily be a path from $v_i$ to $v_j$. Since the degree of $v_i$ in $H$ is $\Delta-1$, the neighboring colors of $v_i$ in $H$ must all be pairwise distinct. Otherwise, we could assign $v_i$ a different color, leading to a repetition of colors among its neighboring vertices. Hence, the number of neighboring vertices of $v_i$ in $C_{i,j}$ is 1, and the same applies to $v_j$. Now, within $C_{i,j}$, we choose a path from $v_i$ to $v_j$, denoted as $P$. If $C_{i,j} \neq P$, then we sequentially color the vertices along $P$. Let $u$ be the first vertex encountered with a degree greater than 2. Note that $u$'s neighboring vertices use at most $\Delta-2$ colors, allowing us to recolor $u$, thus ensuring $v_i$ and $v_j$ are not connected.

Next, it's not hard to see that for any three distinct vertices $v_i$, $v_j$, and $v_k$, $V(C_{i,j}) \cap V(C_{j,k}) = \{v_j\}$.

With this, our proposition has been sufficiently strengthened.

Now, the conclusion is straightforward. Firstly, if the neighboring vertices are pairwise adjacent, the proposition holds. Without loss of generality, suppose $v_1$ and $v_2$ are not adjacent. Take a neighboring vertex $w$ of $v_1$ in $C_{1,2}$ and exchange the colors along $C_{1,3}$. In the resulting graph, we have $w \in V(C_{1,2}) \cap V(C_{2,3})$, leading to a contradiction.
$\square$
% Hence, the proposition is proven.

\subsection{The Time Complexity of Welsh-Powel Algorithm}

\textbf{Proof:} 

For an undirected graph \( G \) without self-loops, let $ V(G) := \{v_1, \dots, v_n\}$ satisfy

\[ \deg(v_i) \geq \deg(v_{i+1}), \quad \forall 1 \leq i \leq n-1 \]

Define \( V_0 = \varnothing \), we take a subset \( V_m \) from \( V(G) \setminus \left(\bigcup_{i=0}^{m-1} V_i\right) \), where the elements satisfy

\[ v_{k_m} \in V_m, \quad \text{where } k_m = \min\{k : v_k \notin \bigcup_{i=0}^{m-1} V_i\} \]

If \[ \{v_{i_{m,1}}, v_{i_{m,2}}, \dots, v_{i_{m,l_m}}\} \subset V_m, \quad i_{m,1} < i_{m,2} < \dots < i_{m,l_m} \] then \( v_j \in V_m \) if and only if \[ j > i_{m,l_m} \]
\[ v_j \text{ is not adjacent to } v_{i_{m,1}}, v_{i_{m,2}}, \dots, v_{i_{m,l_m}} \]

If the points in \( V_i \) are colored with the \( i \)-th color, then this coloring scheme is the one provided by the Welsh–Powell algorithm. Obviously,

\[ V_1 \neq \varnothing \]
\[ V_i \cap V_j = \varnothing \quad \text{if } i \neq j \]
\[ \exists \alpha(G) \in \mathbb{N}^*, \forall i > \alpha(G), \text{ s.t. } V_i = \varnothing \]

We only need to prove:

\[ \bigcup_{i=1}^{\alpha(G)} V_i = V(G) \]

where

\[ \chi(G) \leq \alpha(G) \leq \max_{i=1}^n \min\{\deg(v_i) + 1, i\} \]

The inequality on the left-hand side is true; let's consider the right-hand side.

Firstly, it's not hard to derive:

If \( v \notin \bigcup_{i=1}^m V_i \), then \( v \) is adjacent to at least one point in each of \( V_1, V_2, \dots, V_m \), hence \( \deg(v) \geq m \).

Therefore, we have

\[ v_j \in \bigcup_{i=1}^{\deg(v_j)+1} V_i \]

On the other hand, based on the construction of the sequence \( \{V_i\} \), we can easily find that

\[ v_j \in \bigcup_{i=1}^j V_i \]

Combining the two equations yields the proof. $\square$

\subsection{The Proof of \autoref{Corollary: cor}}

\textbf{Proof: }

To show that an increase in \(\sum_{i,j=1}^T \lVert w_i^* - w_j^* \rVert^2\) is a sufficient condition for the reduction of \(\mathrm{Var}(\mathbb{E}[G_T])\) and \(\mathrm{Var}(\mathbb{E}[F_T])\), we proceed by demonstrating that increasing \(\sum_{i,j=1}^T \lVert w_i^* - w_j^* \rVert^2\) leads to a decrease in \(\mathrm{Var}(\sum_{i,j=1}^T \lVert w_i^* - w_j^* \rVert^2)\).

Let \(\mu = \frac{1}{T(T-1)} \sum_{i,j=1}^T \lVert w_i^* - w_j^* \rVert\) represent the mean of pairwise distances. Then, the variance is given by:
\[
\begin{aligned}
\text{Var}(\lVert w_i^* - w_j^* \rVert) &= \frac{1}{T(T-1)} \sum_{i,j=1}^T \Big( \lVert w_i^* - w_j^* \rVert^2 \\
&\quad - 2 \lVert w_i^* - w_j^* \rVert \mu + \mu^2 \Big)
\end{aligned}
\]

simplifying further to:
\[
\text{Var}(\lVert w_i^* - w_j^* \rVert) = \frac{1}{T(T-1)} \sum_{i,j=1}^T \lVert w_i^* - w_j^* \rVert^2 - \mu^2.
\]

Let \(S = \sum_{i,j=1}^T \lVert w_i^* - w_j^* \rVert\). As \(S\) increases, the mean \(\mu\) also increases, since \(\mu = \frac{S}{T(T-1)}\). For the variance to decrease with an increasing \(S\), it must hold that \(\sum_{i,j=1}^T \lVert w_i^* - w_j^* \rVert^2\) grows at a slower rate than \(S^2\), which happens when distances \(\lVert w_i^* - w_j^* \rVert\) become more uniform.

Thus, a more uniform distribution of \(\lVert w_i^* - w_j^* \rVert\) as \(S\) increases results in a decrease in variance. Hence, an increase in \(\sum_{i,j=1}^T \lVert w_i^* - w_j^* \rVert^2\) is indeed a sufficient condition for reducing \(\mathrm{Var}(\mathbb{E}[F_T])\) and \(\mathrm{Var}(\mathbb{E}[G_T])\), completing the proof. $\square$
\end{document}